\documentclass[runningheads]{llncs}
\usepackage{graphicx}
\usepackage{subfiles}
\usepackage{csvsimple}
\usepackage{float}
\usepackage{comment}
\usepackage{color}
\usepackage{cite}
\usepackage{url}
\usepackage{multirow}
\usepackage{hhline}
\usepackage{caption}
\usepackage{subcaption}
\usepackage[linesnumbered,ruled,vlined]{algorithm2e}
\usepackage{mathtools,xparse}
\usepackage{amsmath}
\usepackage[title]{appendix}
\usepackage{tabularx}
\usepackage{array}
\newcolumntype{?}{!{\vrule width 1pt}}

\begin{document}
\title{Evaluating Explanation Methods for Multivariate Time Series Classification }

\titlerunning{Explaining Multivariate TSC}
%

\author{Davide Italo Serramazza \and Thu Trang Nguyen \and Thach Le Nguyen \and Georgiana Ifrim}
\authorrunning{Serramazza et al.} 
%
\institute{School of Computer Science, University College Dublin, Ireland \\
\email{\{davide.serramazza,thu.nguyen\}@ucdconnect.ie}\\
\email{\{thach.lenguyen,georgiana.ifrim\}@ucd.ie}}
%
\maketitle   
%
\begin{abstract}


Multivariate time series classification is an important computational task arising in applications where data is recorded over time and over multiple channels. For example, a smartwatch can record the acceleration and orientation of a person's motion, and these signals are recorded as multivariate time series. We can classify this data to understand and predict human movement and various properties such as fitness levels. In many applications classification alone is not enough, we often need to classify but also understand what the model learns (e.g., why was a prediction given, based on what information in the data). The main focus of this paper is on analysing and evaluating explanation methods tailored to Multivariate Time Series Classification (MTSC). We focus on saliency-based explanation methods that can point out the most relevant channels and time series points for the classification decision.
We analyse two popular and accurate multivariate time series classifiers, ROCKET and dResNet, as well as two popular explanation methods, SHAP and dCAM. We study these methods on 3 synthetic datasets and 2 real-world datasets and provide a quantitative and qualitative analysis of the explanations provided. 
We find that flattening the multivariate datasets by concatenating the channels works as well as using multivariate classifiers directly and adaptations of SHAP for MTSC work quite well. Additionally, we also find that the popular synthetic datasets we used are not suitable for time series analysis.

\end{abstract}
\section{Introduction}
\label{sec:intro}


Real-world time series data are often multivariate, i.e., data collected over a period of time on different channels. An example is human motion data collected from participants wearing a tri-axial accelerometer on their dominant wrist. The tri-variate data can be examined to identify epilepsy convulsions in everyday life~\cite{epilepsy}. Another example is traffic data where multiple sensors are set up at different locations to measure the traffic occupancy in a city\footnote{\url{https://pems.dot.ca.gov/}}.

While univariate time series have been the main research focus, there is a steadily growing interest in multivariate time series (MTS), in particular for the classification task (MTSC). The release of the MTSC benchmark \cite{bagnall2018uea}, a collaborative effort by researchers from multiple institutions, is an important milestone that has accelerated studies of MTSC methods.   

Explainable AI is another important topic due to the explosion of interest in complex machine learning models and deep learning methods. Pioneers in this field have been working mostly on text and image data and, as a result, a number of explanation frameworks including LIME \cite{ribeiro2016should}, DeepLift \cite{li2021deep}, Shapley \cite{Lundberg2017APredictions} have been introduced. The similarity between image and time series data allows such techniques to be adapted to time series models \cite{wang2017time}.
Nevertheless, there are some notable differences between images and time series. Firstly, images are usually represented using RGB encoding and all the 3 channels contain necessary information, while for time series it is common to have channels that do not contribute to, or even hinder, the classification decision.
Secondly, in images there is a lot of homogeneity in the pixel values while moving between pixels belonging to the same objects and a sharp difference when moving between pixels belonging to different objects. In time series, it is less common to find such a strong locality, especially across all the channels. Furthermore, the data magnitude and pre-processing, such as normalisation, are important factors for time series, but less so for images. 

In this work, we focus on methods for explaining MTSC as this is an important open problem that is often as important as the classification itself. In a scenario in which people wear accelerometers on their body while executing a physical exercise, other than classifying the exercise as correctly executed or not, it is also important to provide feedback to users, e.g., an explanation of why the exercise was incorrectly executed by pointing out the relevant data.

\begin{figure} [ht]
    \centering
    \includegraphics [width=0.92\textwidth]{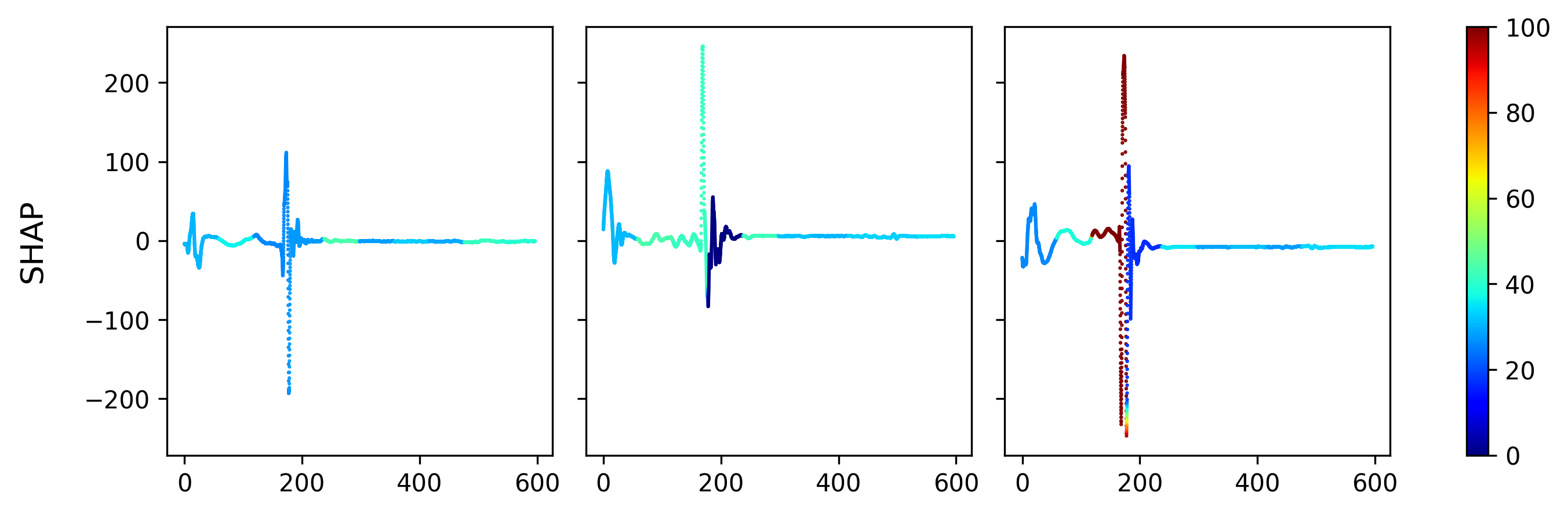}
    \caption{Sample multivariate time series and explanation heat map. The 3 plots show the x, y, z channels for a jump sample.}
    \label{fig:sample_heat_map}
\end{figure}

In this paper, a \textit{multivariate time series explanation} is a 2D saliency map \cite{boniol2022dcam} highlighting the importance of each time series channel and each time point for the classification decision, as illustrated in Figure \ref{fig:sample_heat_map}. A proper MTSC explanation should be able to point out for each channel the relevant time points that may be located at different parts of the time series. For example, CAM \cite{zhou2016learning} was designed for explaining univariate time series thus it can not identify important time points which vary across channels.

In this work we aim to analyse and evaluate a few MTSC explanation methods we found in the literature. Throughout our literature research, the only bespoke MTS explanation methods found are all tailored for deep learning methods (especially CNN), while few others are able to provide a 2D heat map by adapting \textit{univariate time series explanation} to work in a multivariate scenario (most of the time by flattening the dataset and reshaping the 1D heat map into a matrix).

The lack of bespoke multivariate time series explanations, combined with the lack of explanation evaluation methods, is an important gap in the scientific literature. The main aim of this work is to study and evaluate existing MTSC explanation methods in order to start addressing this gap.

\textbf{Our main contributions in this paper are:}

\begin{itemize}
    \item We analyse the literature on saliency-based explanation methods for MTSC and find very few bespoke methods, all of which are designed for deep learning models. 
    Among these, we select dCAM \cite{boniol2022dcam} which extends CAM, a very popular method for time-series and image explanations.
    
    \item We conduct experiments using state-of-the-art multivariate time series classifiers ROCKET \cite{dempster2020rocket} and dResNet \cite{boniol2022dcam} and explanation methods SHAP \cite{NIPS2017_7062} and dCAM \cite{boniol2022dcam}. We study ways to adapt SHAP to work with multivariate time series and compare it to the bespoke MTSC explanation method dCAM.
    
    \item We use 3 synthetic datasets and 2 real-world datasets to compare the classifiers and the explanations. We evaluate the explanations both quantitatively, using the methodology proposed in \cite{nguyen2023amee}, as well as qualitatively. We find that for truly multivariate datasets (i.e., where multiple channels are needed for the correct classification), ROCKET-SHAP works better than dCAM, but is also more computationally expensive. We also find that flattening the datasets by concatenating the channels and using univariate classifiers works as well as using multivariate classifiers directly.
    
\end{itemize}

In the rest of the paper, in Section \ref{sec:rel_works} we discuss prior work addressing the MTSC explanation task. In Section \ref{sec:background} we formally define the problem addressed, the classifiers and the explanation methods used in the experiments. In Section \ref{sec:dataset} we describe the datasets used in our study, in Section \ref{sec:experiments} we describe our experiments and in Section \ref{sec:conclusions} we summarise our main findings.

\section{Related Work}
\label{sec:rel_works}

\textbf{Explanation Methods adapted from Univariate to Multivariate TSC.} Some multivariate time series explanation methods are simple adaptations of methods developed for univariate data.  In \cite{babayev2021interpreting}, the authors  explain the adapted classifiers by applying the timeXplain \cite{mujkanovic2020timexplain} framework on each channel \textit{independently}. The result is a multivariate explanation that highlights  the important segments in each channel of the multivariate sample. Nonetheless, it is arguable whether this approach is appropriate since the explained model(s) (univariate) and the model that needs to be explained (multivariate) are not the same.
Additionally, it is not clear if the accuracy of the channel-wise univariate model is similar or worse than that of the multivariate model, and this is not discussed in the paper.

\noindent\textbf{Bespoke Explanation Methods for MTSC.} Most of the previous  explanation methods designed for MTSC are tailored to deep learning methods, which are not state-of-the-art with regard to classification accuracy.
In \cite{boniol2022dcam}, the authors discussed the drawbacks of the CAM explanation method for MTS data. CAM can only produce a univariate saliency map, thus it is unable to identify the important channels. Features that depend on more than one channel are also not detectable. dCAM, proposed in the same paper, addressed these  limitations by rearranging the input time series with all the permutations of the channels. The paper shows that this technique can be applied to any architecture with a Global Average Pooling layer (GAP)  such as ResNet or InceptionTime. dCAM limitations are discussed by comparing this method with other deep learning explanation methods, as for instance it was shown that dCAM is not the best option when dealing with multivariate datasets that can be classified focusing on just one channel, but there is no comparison against model agnostic methods such as SHAP \cite{Lundberg2017APredictions} or LIME  \cite{ribeiro2016should}.

\noindent\textbf{Evaluation of Explanation Methods for MTSC.} 
While explanation methods for MTSC are few, works on evaluating such methods are even fewer. For univariate time series, several approaches have been proposed to compare explanation methods from different angles. The work in  \cite{Ismail2020BenchmarkingPredictions, Crabbe2021ExplainingMasks} benchmarks the methods with controllable synthetic datasets. The work of \cite{Guidotti2021EvaluatingTruth} attempted to extract "ground-truth" explanations with a white-box classifier. The "ground-truth" explanation is then used to evaluate post-hoc explanations. AMEE \cite{nguyen2023amee} is a recent framework to quantitatively compare explanation methods on a dataset by perturbing input time series and measuring the impact on the classification accuracy of several classifiers. For multivariate time series, recently \cite{naturetsxai} designed an evaluation framework that is also based on the idea of perturbation, but the work is only limited to evaluating deep learning classifiers and associated explanations. The paper also proposed a synthetic multivariate time series dataset to benchmark explanation methods.

\section{Background}
\label{sec:background}

A multivariate time series $X$ can be represented as a $d \times L$ matrix, where the $d$ rows are also called channels and the $L$ columns store the values associated with each channel at every time point. Hence $X_i^j$ is the value of the time series at time point $i$ and channel $j$, with $0 \le i < L$ and $0  \le j < d$. We also refer to $X^j$ as the univariate time series at channel $j$, therefore $X$ can be written as $X=[ X^0,  X^1,  \dots,  X^{d-1} ]$.


An explanation of a time series $X$ is a saliency map $W$ that provides an importance weight for each data point (at every time point $i$ and every channel $j$) in the time series. Hence the saliency map can also be represented by a $d \times L$ matrix. 
A common visualisation method for the saliency map is a heat map where more important data points are highlighted with warmer colours.




An explanation method for MTSC is a method that, given the input MTS, can produce a saliency map highlighting the relevance of each time point to the classifier decision. Intrinsically explainable models such as Ridge Classifier can also be an explanation method while black-box models such as RestNet (dResNet) and ROCKET need a post-hoc explanation method.


In our experiments we compare three different classifiers and explanation methods: ROCKET \cite{dempster2020rocket} coupled with SHAP \cite{NIPS2017_7062}, dResNet coupled with dCAM \cite{boniol2022dcam} and the Ridge Classifier \cite{hoerl1970ridge} which is an intrinsically explainable model. We also use a random explanation (a matrix of random  weights) as a sanity check.

\subsection{Classification Methods}
\label{sec:models}

The first classifier we used is \textbf{ROCKET} \cite{dempster2020rocket} which was originally designed for UTS, but also has an adaptation for MTS: it applies a large set of random convolution kernels to the time series in order to transform it into tabular data with $20,000$ features. It introduced some key concepts such as dilation, proportion of positive values (PPV), etc., starting an algorithm family in which recent members such as Minirocket \cite{dempster2021minirocket}, MultiRocket \cite{tan2022multirocket} are improvements of the original idea.
All the hyper-parameters for ROCKET were learned from the UCR archive. The authors selected 40 random datasets from the archive and used them as the development set to find the best values for the hyper-parameters. Finally, all the kernel weights are sampled from a distribution $\mathcal{N} (0,1)$. After the transformation step, the authors use classic linear classifiers Ridge or Logistic Regression.

The second classifier is \textbf{dResNet} \cite{boniol2022dcam} which is a variation of ResNet \cite{he2015deep}. This last one, originally designed for image classification, was used for the first time in TSC in \cite{wang2017time}. It introduced the key concept of \textit{shortcut connections} to mitigate the gradient vanishing problem. The main architecture of the network is composed of three consecutive blocks which in turn contain three different convolutional layers. These three blocks are followed by a GAP layer and a softmax layer for classification. 
\\ The dResNet version uses the same architecture with two differences specifically designed to work alongside dCAM. Firstly, for a multivariate time series $X$ with $d$ channels, i.e., a matrix $X=[ X^0,  X^1,  \dots,  X^{d-1} ]$, the input $C(X)$ of the network will be a 3D tensor:

\begin{equation*}
C(X)=
\begin{bmatrix}
X^{d-1}  & X^0  & \dots  & X^{d-3}  & X^{d-2} \\ 
\vdots  & \vdots  & \vdots  & \vdots  & \vdots \\
X^1  & X^2  & \dots   & X^{d-1}  & X^0 \\
X^0  & X^1  & \dots  & X^{d-2}   & X^{d-1}  \\
\end{bmatrix}
\end{equation*}
In other words, the input is turned from a $2D$ matrix into a $3D$ one in which each row contains the $d$ channels in different positions. The second change was to turn the convolution shapes from $1D$ to $2D$ to have the same output shape as ResNet. 
These changes were made so that the network is able to capture patterns depending on multiple channels while still learning on individual channels.

The third model we used is the well-known \textbf{Ridge Classifier} \cite{hoerl1970ridge}, meant to be a baseline in the experiments:
we used the scikit-learn \cite{scikit-learn} package RidgeCV using Cross Validation, leaving the other solver parameters as default. 
This classifier disregards the time ordering in each time series as it  treats each time series as a tabular vector of features.

\subsection{Explanation Methods}
\label{sec:expMethods}
The first explanation method considered in this paper is \textbf{SHAP} \cite{Lundberg2017APredictions} which measures feature importance using Shapley values borrowed from game theory. 
SHAP quantifies the contribution of each feature by examining the differences in the model output when a specific feature is masked, i.e., it is replaced with a specific value and when it is not. SHAP considers every possible masking configuration, thus is computationally expensive. The timeXplain library \cite{mujkanovic2020timexplain} applies SHAP on the UTSC task by dividing the time series into segments, each is treated as a feature. The segmentation exploits locality in time series and significantly reduces the number of features before applying SHAP. As SHAP is a model-agnostic method, it works with any TSC model. We couple it with ROCKET due to its efficiency and accuracy.



The second explanation method (used along dResNet), is \textbf{dCAM} \cite{boniol2022dcam}. It computes \textit{CAM} \cite{zhou2016learning} for each row of the input (described in Section \ref{sec:models}), resulting in a $2D$ matrix $\mathcal{M}$ where all channels are brought back to their original positions to evaluate their contribution. 
Since the network is trained to compute meaningful predictions regardless of the order in which the channels are provided, dCAM computes $k$ different matrices $\mathcal{M}$ each of them obtained by a different random permutation of the channel order: all these $k$ matrices are then averaged into $ \hat{ \mathcal{M} }$.
The final step to retrieve the explanation $W$ consists in filtering out uninformative time points and uninformative channels using respectively the average value of $ \hat{ \mathcal{M}}$ in each channel and the variance of all positions for a single channel. dCAM can tell how important a time point was for the classification by taking the differences in $ \hat{ \mathcal{M}}$  when the time point is present in different positions.



The third explanation method is \textbf{Ridge}. As mentioned before, this method is intrinsically explainable because the explanation weights are the weights learned by the classifier. The model is basically a vector of coefficients for each feature, i.e., data point in the time series.


The final explanation method \textbf{Random} is a baseline that generates the saliency map $W$ by sampling values randomly from a continuous uniform distribution. The idea is that any good explanation method should provide a better explanation than the random one.
\section{Datasets}
\label{sec:dataset}

\begin{figure}
    \centering
    \subfloat[]{
        \includegraphics[width=0.23\textwidth]{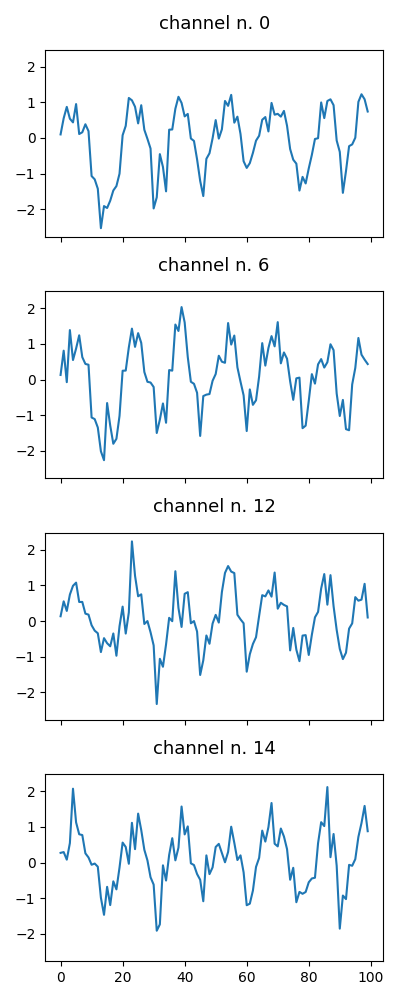}
        \label{fig:synth_plot}} 
    \subfloat[]{
        \includegraphics[width=0.29\textwidth]{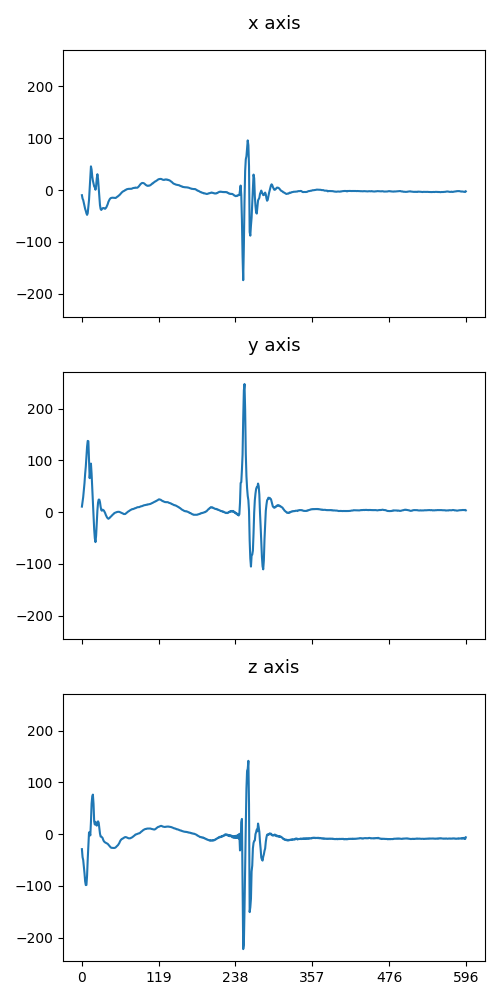} 
        \label{fig:CMJ_plot}}
    \subfloat[]{\includegraphics[width=0.45\textwidth]{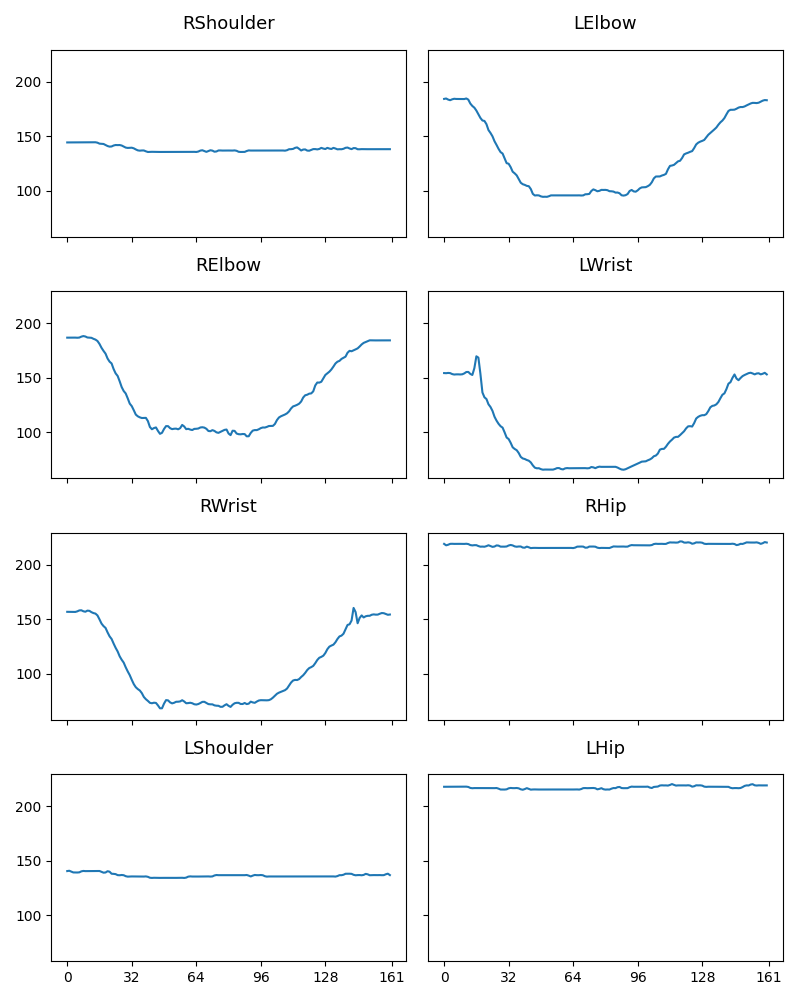}
        \label{fig:MP_plot}}
    \caption{ Sample time series: Fig \ref{fig:synth_plot} PesudoPeriodic negative sample. Fig \ref{fig:CMJ_plot} one instance from CMJ Bend. Fig \ref{fig:MP_plot} one instance from MP Normal.}
    \label{fig:timeSeries_plots}
\end{figure}

We work with 3 synthetic multivariate time series classification datasets and 2 real-world ones. In Figure \ref{fig:timeSeries_plots} we present one sample from one synthetic dataset and one sample each for the real-world datasets.

\subsection{Synthetic Datasets} 
For the synthetic datasets, we use the multivariate time series classification benchmark by Ismail et al. \cite{ismail2020benchmarking}. We generated three different datasets, using the \textit{Pseudo Periodic}, \textit{Gaussian} and \textit{Auto Regressive} distributions. Each has 100 samples in both train and test sets, with $L = 100$ and $d = 20$. The two classes for classification are \textit{positive} and \textit{negative}. The discriminative data points are stationary and within a small box, i.e., $X_i^j$ is discriminative if and only if $10 \le i < 20$ and $0 \le j < 10$. In other words, $50\%$ of the channels and $10\%$ of the time steps are relevant. Overall, only $5\%$ of the time series matter for predicting the class.

\subsection{Real-World Datasets}
The first real-world dataset  is \textbf{Counter Movement Jump} (CMJ)\cite{le2019interpretable}. The data were collected using accelerometer sensors attached on the participants while performing the counter-movement jump exercise. The three classes are: jumps with acceptable form (class 1), with their legs bending during the flight (class 2), and with a stumble upon landing (class 3). The training set has 419 samples while the test set has 179 samples. Each time series has 3 channels ($d=3$) that record the acceleration in $x$, $y$, and $z$ axis. The original data is variable-length thus we resampled every time series to the same length ($L = 596$). From the domain experts, we know that the distinctions between classes are more observable on channel $y$, thus it makes this channel the most important one.


The second real-world dataset is \textbf{Military Press} (MP) \cite{singh:dami23}. To collect the data, 56 participants were asked to perform the Military Press strength-and-conditioning exercise. Each of them completed 10 repetitions in the normal form and another 30  in induced forms, with 10 repetitions each  (simulating 3 types of errors). The time series were extracted from video using the OpenPose library \cite{8765346}.
The dataset has 1452 samples in the training set and 601 in the test set, each time series has 161 time points and 50 channels corresponding to the $x,y$ coordinates of 25 body parts. From the original dataset we have selected 8 channels representing the $y$ coordinates of both left and right Shoulder, Elbow, Wrist and Hip.
This dataset has 4 different classes representing the kind of exercise done, namely  Normal (N), Asymmetrical (A), Reduced Range (R) and Arch (Arch). We know from domain experts that the importance of channels for this dataset is in decending order: Elbows, Wrists, Shoulders, Hips. High accuracy can be obtained only by using the Elbows and Wrists while it is not possible to achieve a high accuracy by only using one channel. We later show experiments both in Section \ref{sec:experiments} and in the Appendix to document this behaviour.
\section{Experiments}
\label{sec:experiments}

In our experiments we aim to understand the strengths and weaknesses of existing methods for explaining multivariate time series classification. As summarised in Table \ref{tab:exp_methods}, we compared one of the bespoke multivariate method found (dResNet), the popular SHAP, which has the downside of being adapted to provide a 2D heatmap, and Ridge as a sanity check baseline. Some different coupling such as ROCKET paired with dCAM or dResNet paired with SHAP are not possible respectively because dCAM can only explain models having a GAP layer and the timeXplain library (used for ROCKET-SHAP concatenated) is implemented only for 1D-vector instances (univariate time series).

\begin{table} [ht]
    
    \centering
    \begin{tabular}{|c|c|c|}
        \hline
         Classifier & Explanation Method & MTS Approach \\
        \hline
        dResNet & dCAM & Bespoke MTSC  \\
        ROCKET & SHAP & Concatenated   \\
        
        ROCKET & SHAP & Channel by Channel   \\
        Ridge Classifier & Ridge Classifier & Concatenated   \\
        n/a & Random & n/a \\       
        \hline
    \end{tabular}
    \caption{Summary of the explanation methods tested in this paper.}
    \label{tab:exp_methods}
    \vspace{-8mm}
\end{table}

\begin{figure} [ht]
    \centering
    \subfloat[\label{fig:shap_chBych}]{\includegraphics[width=0.46\textwidth]{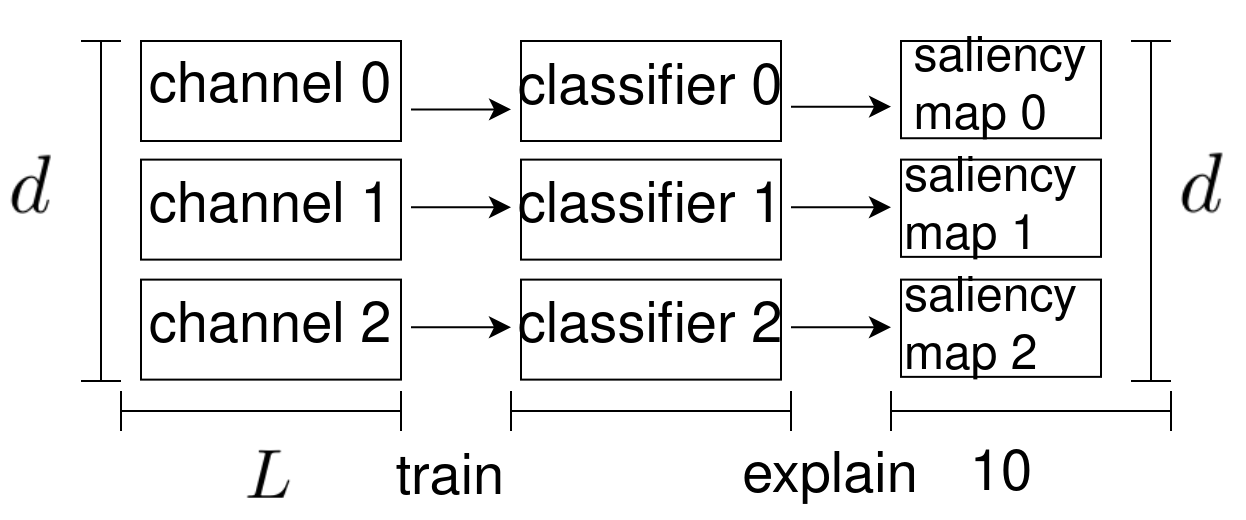} }
    \hspace{0.05\textwidth}
    \subfloat[\label{fig:shap_flatten}]{\includegraphics[width=0.46\textwidth]{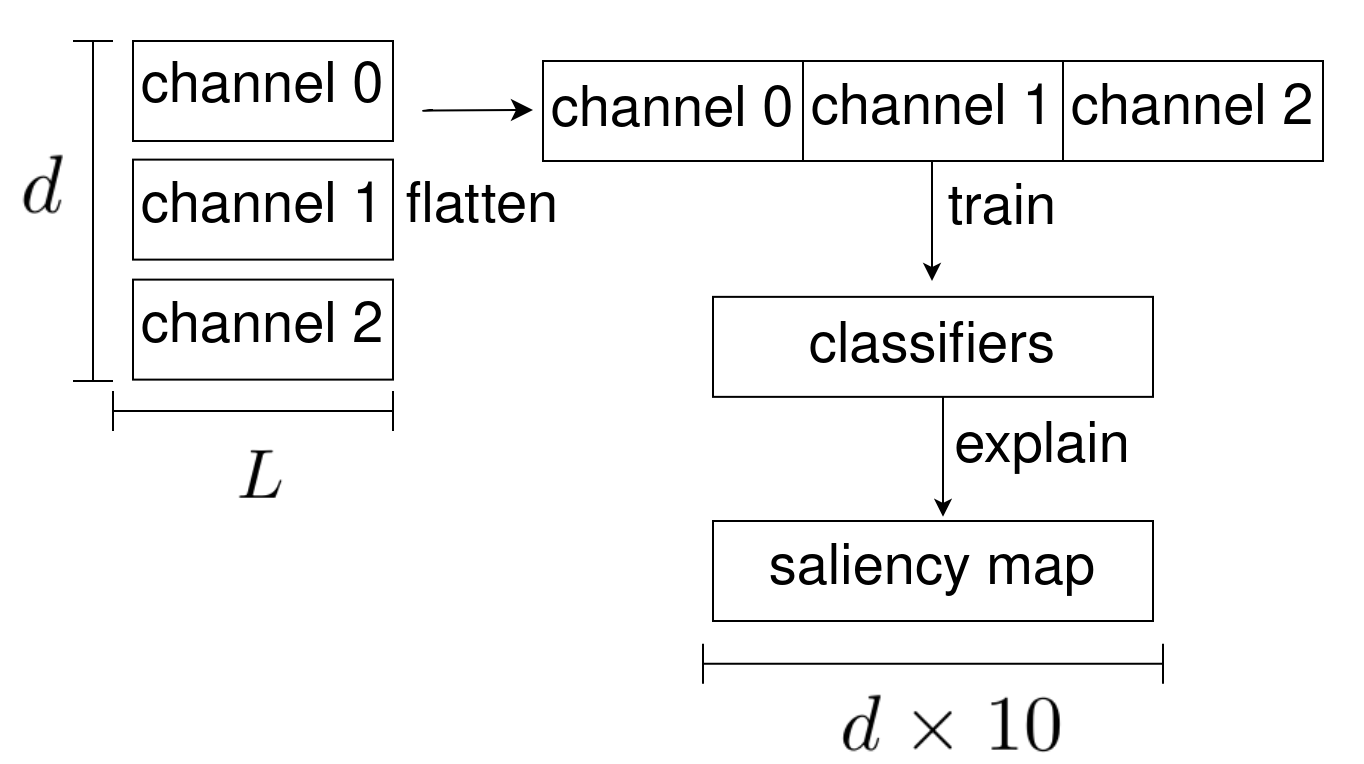}  }
    \caption{Strategies to use the timeXplain library in a multivariate scenario, for $d=3$. In Fig \ref{fig:shap_chBych}, a classifier is trained for each channel: for explaining each classifier, $d$ heat maps of length $10$ are produced: stacking these vectors together results in a matrix of dimension $d \times 10$ . In Fig \ref{fig:shap_flatten} the time series are concatenated and one single classifier is trained. We explain the classifier using a number of segments $d \times 10$ and reshape the resulting vector into a 2D matrix having the same shape as in the previous case.}
    \label{fig:different_shap}
\end{figure}

To make the timeXplain library work with MTS, we apply the following two strategies (Figure \ref{fig:different_shap}): (1) \textbf{Concatenated}: Concatenating all the channels to a single univariate time series. As a result, the output saliency map is also univariate and thus needs to be reshaped. (2) \textbf{Channel by Channel}: Train and explain one model for each channel independently. The MTSC model in this case is an ensemble of per-channel UTSC models.

For SHAP-channel-by-channel, we assign the number of segments to 10 while, for SHAP-concatenated, the number of segments is set to $ d \times 10$.
Since Ridge can only work using univariate datasets, we only used the dataset concatenation strategy for this classifier.
The output of all explanation methods is a saliency map in the form of either $d \times L$ or $d \times 10$ matrix (reshaped if necessary). 


It is important to note that we have only one bespoke method for multivariate time series, dCAM, that computes a saliency map of the same shape as the original time series instance.

All the experiments were done using a machine with 32GB RAM, Intel i7-12700H CPU and an Nvidia GeForce RTX 350 Ti GPU (the GPU was used only for dResNet/dCAM).
All the code used to perform the experiments is available on a Github repository\footnote{\url{https://github.com/mlgig/Evaluating-Explanation-Methods-for-MTSC}}.

\subsection{Classification Accuracy Analysis}
\label{sec:accuracy}
Before diving into the explanations, we first take a look at the accuracy of the classifiers used for producing the explanations. All the classifiers listed in Table \ref{tab:original_accuracies} were trained 5 different times (for ROCKET we also tried to either normalize the data or not). In this Table are reported the most accurate ones i.e., the models used in the experiments as well as the accuracy for the univariate concatenated datasets.


Having a look at the table we can notice that all the times both ROCKET and dResNet have high accuracy (with some exceptions for the synthetic datasets): this is an important pre-requisite when comparing explanations methods applied to different classifiers as we did. 

We note that RidgeCV does particularly well on the synthetic datasets. 
On Military Press, the multivariate models are more accurate than the univariate ones (on concatednated data). This is expected since it is difficult to achieve a high accuracy with a single channel for this dataset, so this is a trully multivariate dataset. 
Concatenating all the channels for Military Press hurts more dResNet which loses 9 percentage points accuracy, while ROCKET loses only 4. For CMJ, the behaviour is reversed, with univariate models being more accurate than the multivariate ones. 
dResNet has a noticeable 9 percentage points improvement on the concatenated dataset, while ROCKET gains 1 percentage point.
\begin{table} [htb]
    \centering
    \begin{tabular}{|c?c|c|c?c|c|c|}
        \hline
         Classifier/Dataset & PseudoPeriodic & Gaussian & AutoRegressive & CMJ  & MilitaryPress \\
        \hline
        dResNet multivariate & 1.0 & 0.83 & 0.82 & 0.82 & 0.79 \\
        dResNet concatenated & 1.0 & 0.89 & 0.81 & 0.91 & 0.68  \\
        \hline
        ROCKET multivariate & 1.0 & 0.93 & 0.87 & 0.87 & 0.87  \\
        ROCKET concatenated & 1.0 & 0.72 & 0.73 & 0.88 & 0.83  \\
        ROCKET ch-by-ch & 0.99 & 0.72 & 0.95 & 0.85 & 0.65  \\
        \hline
        RidgeCV & 1.0 & 1.0 & 1.0 & 0.44 & 0.61 \\
        \hline
    \end{tabular}
    \caption{Accuracy for the models listed in Table \ref{tab:exp_methods} plus dResNet concatenated and ROCKET multivariate: using this table it is possible to appreciate the differences when using multivariate vs univariate datasets.} 
    \label{tab:original_accuracies}
\end{table}

\subsection{Synthetic Data}
\label{sec:exp_synth}
For the synthetic data, we performed 5-fold  cross-validation to train a Logistic Regression classifier for ROCKET, allowing up to 1000 iterations. For dResNet we used 64 filters, and we trained using the Cross-Entropy Loss and Adam optimizer with a learning rate set to $0.0001$. Finally for RidgeCV we used the standard scikit-learn parameters for cross-validation using 5 folds.
\\Regarding the explanation methods we used $10$ segments for ROCKET concatenated in the channel-by-channel scenario and $200$ segments in the concatenated one; for dCAM the number of permutations to evaluate $k$ was set to $200$ (this is the maximum recommended in \cite{boniol2022dcam}). 

\begin{figure} [ht!]
    \centering
    \includegraphics[width=0.95\textwidth]{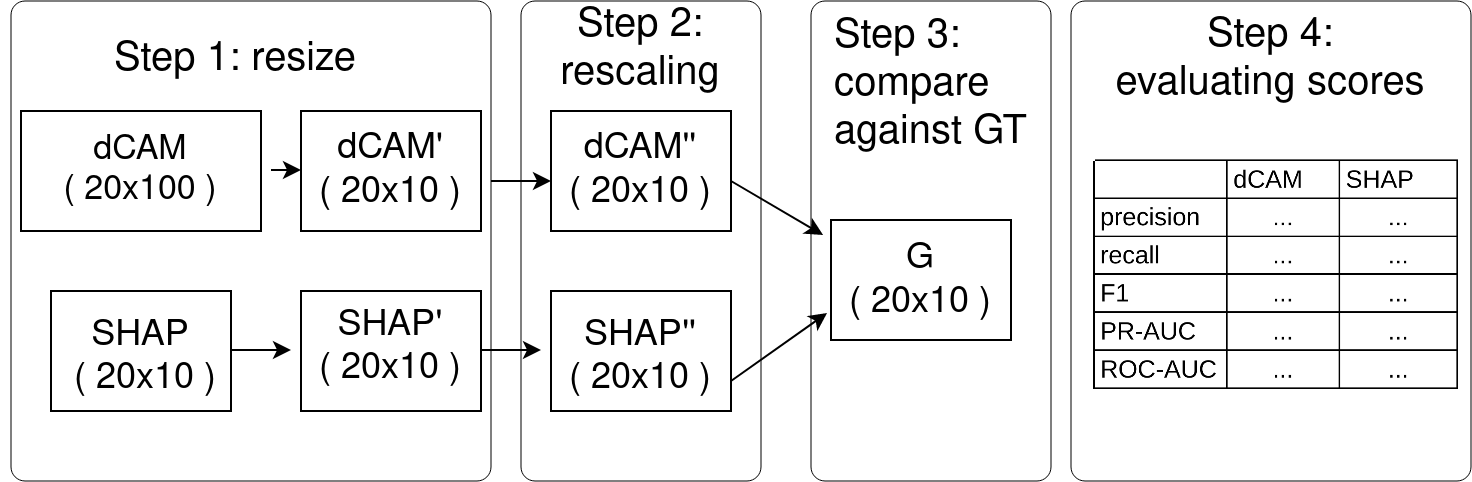}
    \caption{Steps performed in the synthetic data evaluation when comparing dCAM and SHAP. In Step 1,  dCAM is reshaped into $(20,10)$ averaging 10 consecutive elements in each channel, while SHAP is untouched. In Step 2, the reshaped matrices are rescaled in the range $[0,100]$. In Step 3, both the explanations achieved so far are compared against the ground truth matrix $G$ and finally in Step 4 the scores computed in the previous step are evaluated.}
    \label{fig:synthetic_steps}
\end{figure}

The steps done for syntethic data evaluation are illustrated in Figure \ref{fig:synthetic_steps}.
The first step is to reshape all the explanations so that they all have the same dimension. Specifically, the saliency maps we obtained from dCAM and Ridge have a shape of $d \times L=20 \times 100$ while the ones from SHAP concatenated and channel-by-channel have a shape of $d \times \text{n. segments} = 20 \times 10 $. We chose to average 10 consecutive elements for dCAM and Ridge explanation as we empirically verified that all the metrics had slight improvements. The alternative was to repeat 10 times the same item in SHAP explanations. 
\\After this stage, all explanations have the shape of $20 \times 10$.

The second step rescales the explanation weights as they can have different magnitudes among different instances and different methods. First of all, we take the absolute value of each explanation (to also take into consideration variables that have a negative contribution for the classification) and then we rescale  by min-max normalization in the range $[0,100]$.

The third step is to instantiate a ground truth matrix $G$ and compare each explanation against it. For the settings described before, this is a binary matrix having shape $ 20 \times 10 $ (same dimension of explanations after Step 2), all the elements are set to 0 except for the ones in $G_i^j  \text{ with } i=1, 0 \leq j <10$  that are set to $1$. In other words, this is a binary matrix describing whether or not a segment is important for the classification.
\\ To be noted that synthetic dataset parameters such as the number and range of informative time points and channels, and explanation method parameters such as the number of segments were chosen so that the resulting segments are made up either by only informative time points or by only uninformative time points.

The last step is simply to compute the metrics used for the evaluation i.e.,  Precision, Recall, F1-score, PR-AUC and ROC-AUC \cite{boniol2022dcam}.
For Precision and Recall we had to fix a threshold dividing the values considered uninformative from the ones considered informative: we have chosen $50$ as the medium value between $0$ and $100$.
On the other hand PR-AUC and ROC-AUC do not fix any threshold as they average multiple scores using different thresholds into one single value.
All these metrics computed for the 3 synthetic datasets are reported in Table \ref{tab:synth_metrics}.

\begin{table}[htb]
    \centering
    \small
    \begin{tabular}{|c|c|c|c|c|c|c|c|c|}
        \hline
        Dataset  & XAI Method  & Precision & Recall & F1 & PR-AUC & ROC-AUC & Time \\\hline
        
        Pseudo-Periodic & SHAP ch-by-ch & 0.73 & 0.94 & 0.82 & 0.99 & 0.99 & 6.2 h \\
        
        Pseudo-Periodic & SHAP concatenated & 0.92 & 0.66 & 0.77 & 0.99 & 0.99 & 3.5 h \\
        
       Pseudo-Periodic & dCAM & 0.50 & 0.50 & 0.50 & 0.63 & 0.98 & 50 m \\

       Pseudo-Periodic & Ridge & 1.0 & 1.0 & 1.0 & 1.0 & 1.0 & 0 s  \\
        \hline

        Gaussian & SHAP ch-by-ch & 0.88 & 0.63 & 0.73 & 0.91 & 0.99 & 6.2 h   \\
        
        Gaussian & SHAP concatenated & 0.34 & 0.18 & 0.24 & 0.16 & 0.71 & 3.5 hr \\
        
        Gaussian & dCAM & 0.36 & 0.15 & 0.21 & 0.35 & 0.94  & 50 m \\
        
        Gaussian & Ridge & 0.83 & 1.0 & 0.9 & 1.0 & 1.0 & 0 s \\
        \hline

        Auto-Regressive & SHAP ch-by-ch & 0.85 & 0.60 & 0.71 & 0.49 & 0.77 & 6.2 h   \\
        
        Auto-Regressive & SHAP concatenated & 0.27 & 0.13 & 0.18 & 0.29 & 0.57 & 3.5 h\\
        
        Auto-Regressive & dCAM & 0.34 & 0.15 & 0.21 & 0.06 & 0.57  & 50 m \\
        
        Auto-Regressive & Ridge & 1.0 & 1.0 & 1.0 & 1.0 & 1.0 & 0 s \\
        \hline

        All   \ & Random & 0.05 & 0.15 & 0.08 & 0.05 & 0.5 & 0 s\ \\
        \hline

    \end{tabular}
    \caption{ Scores and runtime of each XAI method for synthetic datasets: h stands for hours, m stands for minutes and s stand for seconds; ch-by-ch stands for channel by channel.}
    \label{tab:synth_metrics}
\end{table}

Looking at the table it is possible to note that all the time Ridge has perfect metrics but for Recall (and consequently F1 score) with the Gaussian dataset. These results along with the one provided in Table \ref{tab:original_accuracies} (perfect accuracies of Ridge for the 3 synthetic datasets), are very strong evidence that these commonly used benchmarks are not ideal for time series analyses at least using the parameters described before. We think this is the case due to the way the benchmarks are created, by adding or subtracting a single value to consecutive time points. This means that a simple tabular classifier such as Ridge is enough to perfectly classify these datasets. In conclusion, we recommend against the use of these synthetic benchmarks for analysing time series classification or explanation methods.

Comparing the other method, most of the time SHAP channel by channel is the second best model, while comparing dCAM with SHAP concatenated there is no clear winner as in some metrics the first one has better results while in some others is the opposite.

The two last points to be noted are that some methods have metrics  close to random, especially for Recall, and the time required for computing the explanations is high, taking into account that these are small datasets: 50 minutes for dCAM, 3.5 hours for SHAP concatenated, and more than 6 hours for SHAP channel by channel. 
\subsection{Real-world Data}
\label{sec:realistic}
In this section we used some different hyper-parameters: for dResNet the number of filters is now set to $128$ as we found better classification results, the number of dCAM permutations $k$ was set to $6$ for dCAM (this dataset has 3 channels so the number of possible channel permutations is just 6) while it is still $200$, i.e. the maximum recommended, for MP which has 8 channels.
We set the number of timeXplain segments using the concatenated dataset to $30$ for CMJ and $80$ for MP so that they are still equal to $ d \times 10 $.

Looking at the classifier accuracy in Table \ref{tab:original_accuracies} we notice how for the two real-world datasets, the accuracies achieved by dResNet and ROCKET are comparable or even better when using the concatenated dataset versions. This means that analysing the explanation methods for MTSC by turning the multivariate problems into univariate ones could be useful.  

The close accuracy between original multivariate and concatenated univariate datasets can arise some questions whether these datasets are truly multivariate (i.e., the necessary information for correct classification is spread among different channels). This seems to be the case for Military Press, but less so for CMJ.
We plan to investigate further this point in future work.

In this work we decided to use the concatenated datasets and the methodology developed by \cite{nguyen2023amee} to evaluate the explanation methods. For the case of dCAM which produces a matrix as an explanation, we flatten the matrix to a vector by concatenating the rows and using it as any other univariate explanation. So dCAM is obtained in a truly multivariate setting (dResNet is a multivariate classifier and dCAM a multivariate explanation), but reshaped to look like a univariate explanation. The explanations obtained from SHAP and Ridge, on the other hand, are univariate explanations obtained  by first concatenating the channels and then running univariate classifiers.

For the real-world datasets we do not have precise explanation ground truth as for the synthetic datasets, but we do have domain knowledge about which channels and parts of the time series are important for the classification.
\\ Finally in this section we didn't include SHAP channel by channel in the MP dataset experiment as the accuracy is low (Table \ref{tab:original_accuracies}) therefore it does not make sense to derive an explanation. 

\subsubsection{Evaluation of Explanation Methods.} We apply AMEE \cite{nguyen2023amee}, an explanation evaluation method for the univariate time series classification task, on the CMJ and MP univariate datasets obtained through concatenating all channels. This method aims to measure the faithfulness of an explanation by estimating its impact on a set of independent classifiers (the \textit{referee classifiers}). If an explanation correctly points out the important areas of a univariate time series, perturbation of such areas will lead to a drop in accuracies of the referee classifiers. The faithfulness of the explanation is then calculated using the Area Under the Curve (AUC) of the accuracy drop curves of each of the referee classifiers. AMEE is designed to be robust by employing various perturbation strategies (i.e. how an important area is perturbed and replaced with a new value) and a diverse set of high-performing referee classifiers. The main idea is that masking away important parts of the data as pointed out by the explanation, should affect a set of high performing classifiers leading to a drop in accuracy across the board.

\begin{table}[htb]
    \centering
    \begin{tabular}{|c|c|c|c|}
        \hline
        Dataset  &  MrSEQL & ROCKET & WEASEL 2.0 \\
        \hline
        CMJ-concat & 0.76 & 0.88  & 0.92  \\
        \hline
        MP-concat & 0.82 & 0.84 & 0.80 \\
        \hline
    \end{tabular}
    \caption{Accuracy of referee classifiers for the AMEE evaluation of explanation methods for univariate time series classification.}
    \label{tab:accuracies-ref}
\end{table}

For our task, we use the default perturbation strategies with three classifiers included in the standard referees set:  MrSEQL \cite{le2019interpretable} WEASEL 2.0 \cite{Schafer2023WEASELClassification} and ROCKET (for more information and results about this methodology we invite the readers to have a look to the original publication \cite{nguyen2023amee}). Table \ref{tab:accuracies-ref} shows the accuracy of these referee classifiers on the evaluated datasets. 

The result of the explanation evaluation is presented in Table \ref{tab:AMEE} as well as the methodology and the evaluation running time. The methodology running time is dependent on the number of both perturbation strategies and employed referees. It is specific to our choice of the three mentioned referees and four perturbation types using Mean and Gaussian sample from both time-point dependent (local) and time-point independent (global) statistics of the test samples. 
Looking at the second one (time for running the explanation methods) we notice the high SHAP computational complexity: this was the main reason why we used only 2 real-world datasets for the experiments. We focused on human motion data because in this case we can rely on domain expertise. 

From the quantitative evaluation with AMEE, we note that for the CMJ dataset, SHAP concat is the best method, although it is close to a random explanation. dCAM ranks third for this dataset. We note that this dataset is quite noisy due to quiet parts after the jump, and this could explain why SHAP and Random are so close in ranking.
\\For the MP dataset, SHAP concatenated is by far the best method, significantly better than dCAM, as well as Random and Ridge. This is an interesting finding considering that dCAM was proposed to deal with datasets where there are clear dependencies between channels, but for MP this method does not seem to perform so well.
\\We supplement the quantitative ranking with a more detailed qualitative analysis in the Appendix. In short we find that for CMJ, the importance rankings of channels given by SHAP concat and dCAM are the same, while for MP, SHAP provides a ranking more in line with domain knowledge, while dCAM places the least informative channels at the top of the ranking.

\begin{table}[htb]
\centering
\begin{tabular}{|l|l|c|c|c|c|}
\hline
Dataset                     & XAI method    & \multicolumn{1}{l|}{Explanation Power} & \multicolumn{1}{l|}{Rank} & \multicolumn{1}{l|}{Evaluation Time}  & \multicolumn{1}{l|}{Explanation Time}  \\ \hline
\multirow{5}{*}{CMJ-concat} & SHAP concat   & 1.00                                   & 1                         & 2h               & 7.15h                                         \\ \cline{2-6} 
                            & Random        & 0.99                                   & 2                         & 2h               & 0s                \\ \cline{2-6} 
                            & dCAM          & 0.39                                   & 3                         & 2h               & 30s                 \\ \cline{2-6} 
                            & SHAP ch-by-ch & 0.05                                   & 4                         & 2h               & 7.5h               \\ \cline{2-6} 
                            & Ridge         & 0.0                                    & 5                         & 2h               & 0s                 \\ \hline
\multirow{4}{*}{MP-concat}  & SHAP concat   & 1.00                                   & 1                         & 4.8h             & 24h                                            \\ \cline{2-6} 
                            & dCAM          & 0.33                                   & 2                         & 4.8h             & 15m               \\ \cline{2-6} 
                            & Random        & 0.07                                   & 3                         & 4.8h              & 0s              \\ \cline{2-6} 
                            & Ridge         & 0.0                                    & 4                         & 4.8h              & 0s              \\ \hline
\end{tabular}
\caption{Results of AMEE to rank XAI methods on CMJ and MP datasets concatenated. }
\label{tab:AMEE}
\end{table}

\section{Conclusion}
\label{sec:conclusions}

In this paper we have investigated explanation methods for MTSC. We studied two very popular explanation methods, dCAM and SHAP, and have provided a quantitative and qualitative analysis of their behavior on synthetic as well as real-world datasets. We found that adaptations of SHAP for MTSC work quite well, and they outperform the recent bespoke MTSC explanation method dCAM. We have also pointed out that a very popular synthetic MTSC benchmark does not seem suitable for MTSC evaluation, since a simple Ridge classifier outperforms all other methods both in classification accuracy and in explanation quality. Finally, while SHAP seems to work effectively to point out important time series channels and time points, we highlighted the time required to run SHAP and pointed out the open problem of excessive time requirements for this method. In future work we plan to investigate the computation time for SHAP, as well as other frameworks for evaluating bespoke explanation methods for MTSC.



\section*{Acknowledgments}
This publication has emanated from research supported in part by a grant from Science Foundation Ireland under Grant number 18/CRT/6183. For the purpose of Open Access, the author has applied a CC BY public copyright licence to any Author Accepted Manuscript version arising from this submission.

%
%
%
%

\bibliographystyle{splncs04}
\bibliography{ref.bib}
\newpage
\begin{appendix}
\section{Supplementary Material}

\begin{table}[ht]
    \begin{subtable}[ht]{0.3\textwidth}
        \centering
        \begin{tabular}{|c|c|}
            \hline
            channel & importance \\
            \hline
            LElbow & $0.60$ \\
            RElbow & $0.59$ \\
            RWrist & $0.58$ \\
            LWrist & $0.57$ \\
            LShoulder & $0.52$ \\
            RShoulder & $0.49$ \\
            LHip & $0.39$ \\
            RHip & $0.36$ \\
            \hline
       \end{tabular}
       \caption{Rocket ranking}
       \label{tab:rocket_MP_colRank}
    \end{subtable}
    \hfill
    \begin{subtable}[ht]{0.3\textwidth}
        \centering
        \begin{tabular}{|c | c |}
            \hline
            channel & importance \\
            \hline
            LHip & $1.0$ \\
            RHip & $0.99$ \\
            RWrist & $0.74$ \\
            LWrist & $0.73$ \\
            RElbow & $0.54$ \\
            LElbow & $0.53 $ \\
            RShoulder & $0.50 $ \\
            LShoulder & $0.50$ \\
            \hline
        \end{tabular}
        \caption{dCAM ranking}
        \label{tab:dCAM_MP_colRank}
     \end{subtable}
    \hfill
    \begin{subtable}[ht]{0.3\textwidth}
        \centering
        \begin{tabular}{|c | c |}
            \hline
            channel & importance \\
            \hline
            RWrist & $1.0$ \\ 
            LWrist & $0.98$ \\ 
            LElbow & $0.93$ \\ 
            RElbow & $0.86$ \\ 
            LShoulder & $0.82$ \\ 
            RShoulder & $0.80$ \\ 
            RHip & $0.77$ \\ 
            LHip & $0.76$ \\
            \hline
        \end{tabular}
        \caption{SHAP ranking}
        \label{tab:SHAP_MP_colRank}
     \end{subtable}
     
     \caption{Ranking of MP columns using different methods. In Table \ref{tab:rocket_MP_colRank} ordered accuracy of single-channel classifiers in ROCKET channel-by-channel scenario: we take this as a channel importance baseline. In Table \ref{tab:dCAM_MP_colRank} and \ref{tab:SHAP_MP_colRank} respectively dCAM and SHAP ranking achieved by averaging all time points in the single channels among all the time series (for readability the values were rescaled such that the most important channel has a value $1.0$).  
     \\Table \ref{tab:SHAP_MP_colRank} is closer  to  \ref{tab:rocket_MP_colRank} than  Table \ref{tab:dCAM_MP_colRank}. The major difference is in the RHip and LHip ranking: while SHAP places them as the least important ones, agreeing with Rocket, dCAM ranks them as the most important channels. }
     \label{tab:MP_channel rankings}
\end{table}

\begin{table}[ht]
    \centering
    \begin{tabular}{|c|c|c|c|c|c|c|c|c|}
        \hline
         method & LElbow & RElbow & RWrist & LWrist &  LShoulder &  RShoulder & RHip & LHip \\ 
         \hline 
         ROCKET & 1 & 2 &  3 & 4 & 5 & 6 & 7 & 8 \\
         \hline
         dCAM & 6 & 5 & 3 & 4 & 8 & 7 & 2 & 1 \\
         SHAP & 3 & 4 & 1 & 2 & 5 & 6 & 7 & 8 \\
        \hline
    \end{tabular}
    \caption{MP channels importance: same analysis as in Table \ref{tab:MP_channel rankings} but showing the ranking rather than the raw values. }
    \label{tab:MP_channel_importance}
\end{table}

\begin{table}[htb]
    \centering
    \begin{tabular}{|c|c|c|c|}
        \hline
         Method & y & z & x  \\ 
         \hline
         ROCKET & 0.85 & 0.81 & 0.79 \\
         \hline
         dCAM & 1.0 & 0.92 &  0.89 \\
         SHAP & 1.0 & 0.85 & 0.83 \\
        \hline
    \end{tabular}
    \caption{Ranking of CMJ columns using different methods. In this case all the methods agree.}
    \label{tab:CMJ_channel_importance}
\end{table}

\begin{figure}
    \centering
    \includegraphics[width=0.92\textwidth]{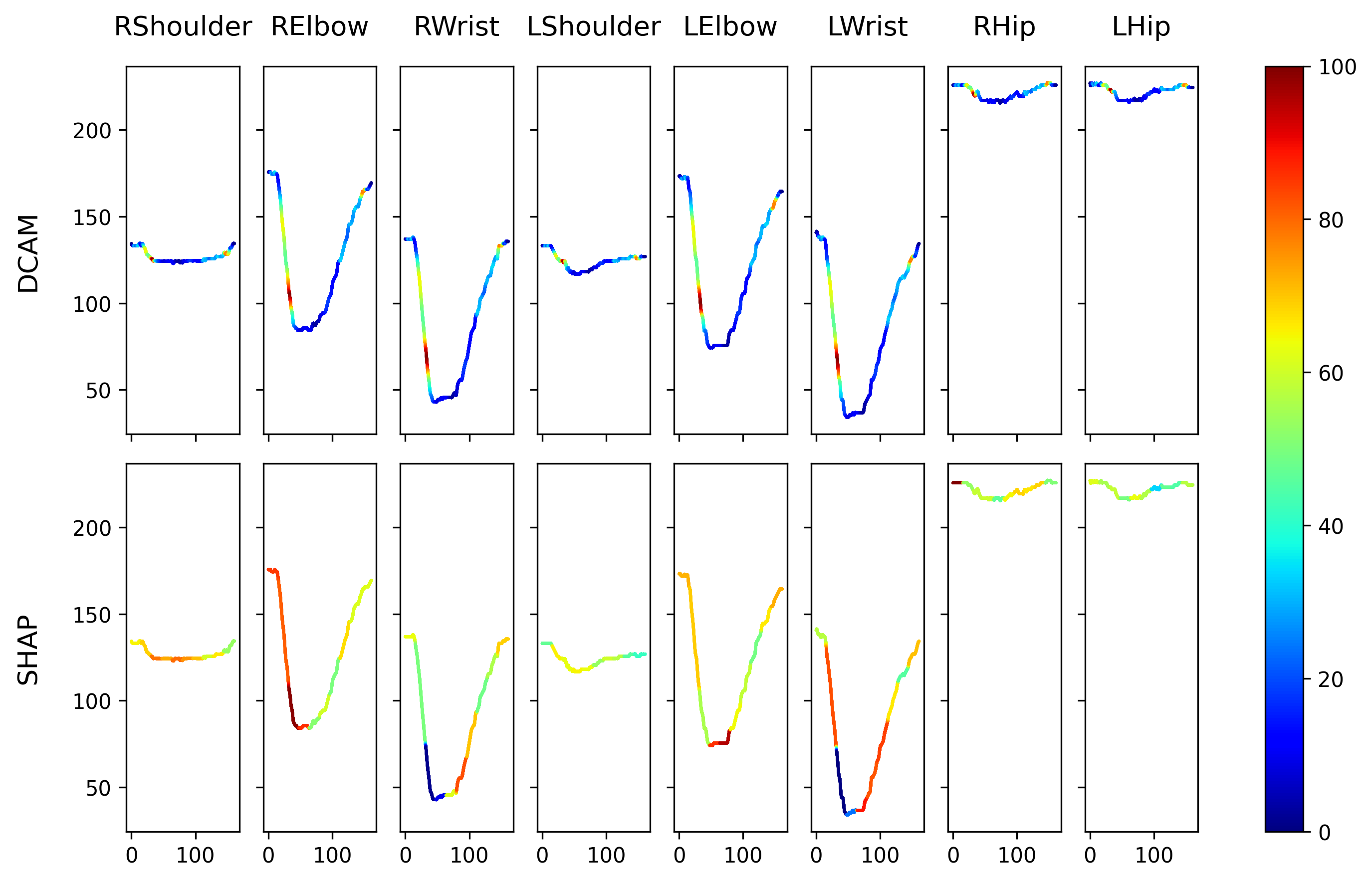}
    \caption{Different saliency maps for the same MP instance correctly classified as Normal: SHAP focuses on LWrist, RElbow, RWrist and LElbow. dCAM has a similar behavior in terms  of channel importance but it focuses on smaller sections: this is just partially explainable due to division in segments by SHAP since sometimes there is no overlapping between highlighted regions in the same channel.}
    \label{fig:MP_n_appendix}
\end{figure}

\begin{figure}
    \centering
    \includegraphics[width=0.92\textwidth]{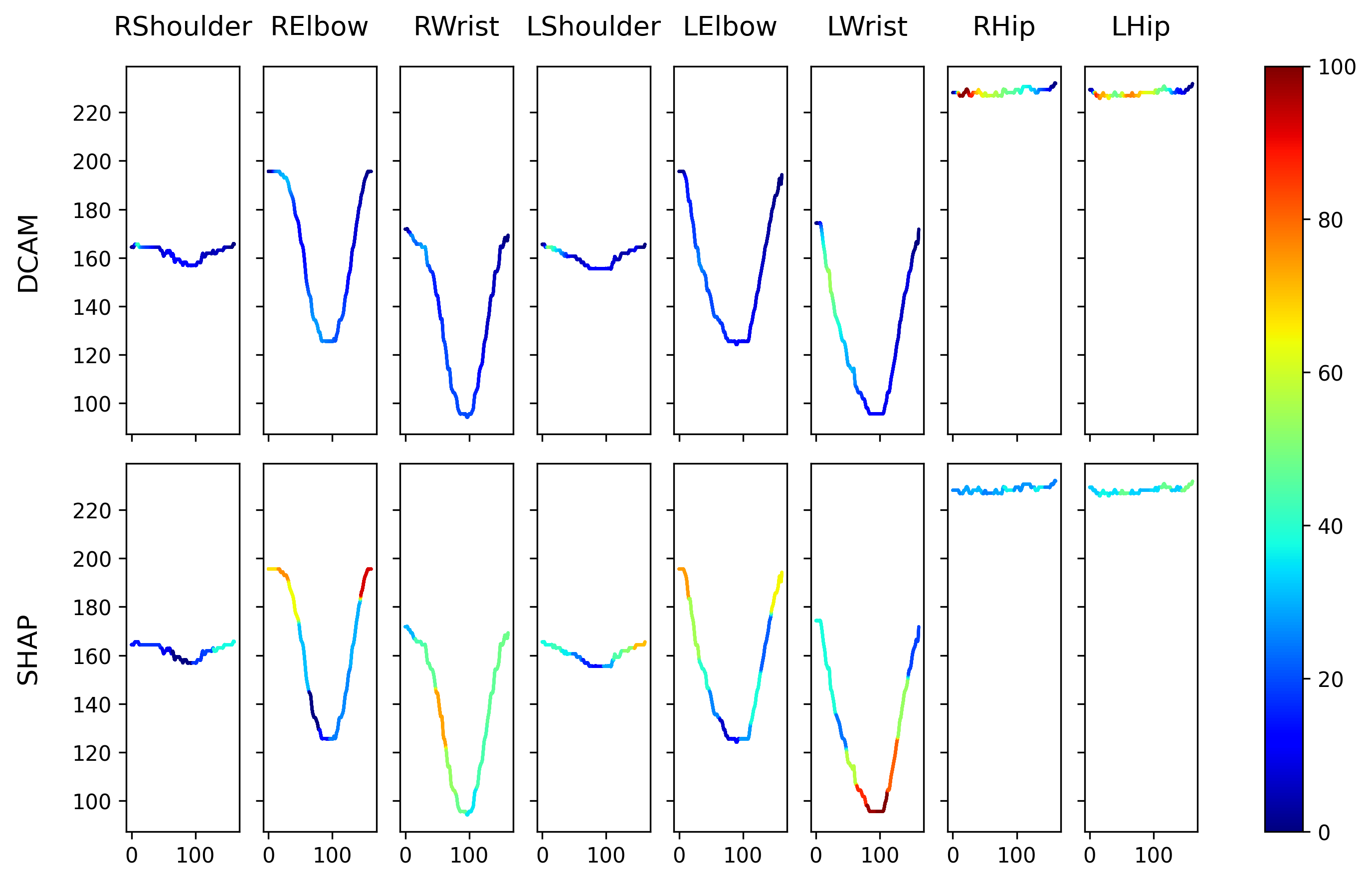}
    \caption{ Different saliency maps for the same MP instance correctly classified as Asymmetric. dCAM focuses on Hips channels while SHAP focuses on Wrists and Elbows in accordance to Table \ref{tab:MP_channel_importance}.}
    \label{fig:MP_a_appendix}
\end{figure}

\begin{figure}
    \centering
    \includegraphics[width=0.95\textwidth]{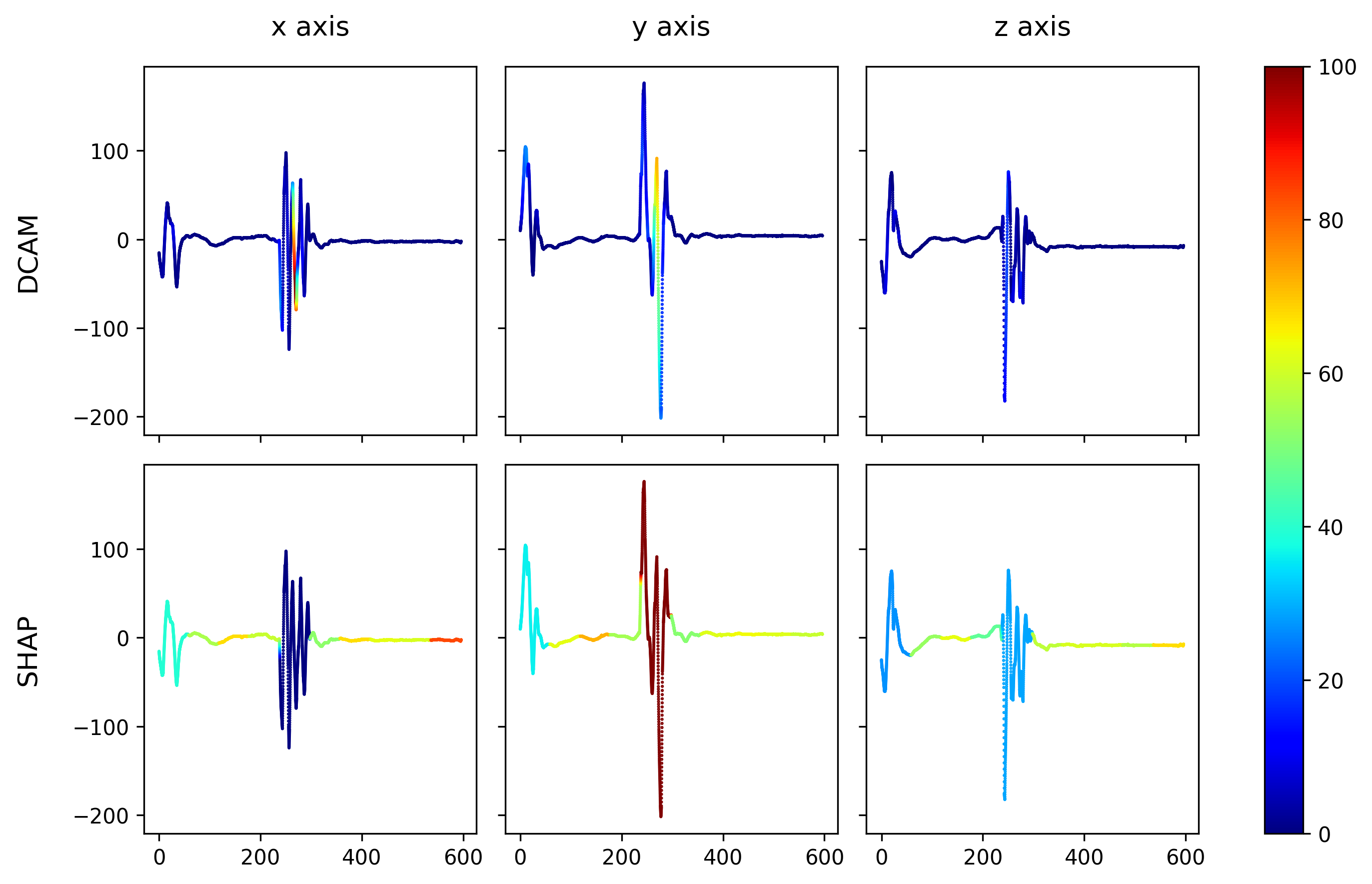}
    \caption{ Different saliency maps for the same CMJ instance correctly classified as Acceptable form: SHAP focuses more on the $y$ axis according to what is shown in Table \ref{tab:CMJ_channel_importance} while dCAM highlights more the $x$ axis.}
    \label{fig:CMJ_acceptable_form_appendix}
\end{figure}

\begin{figure}
    \centering
    \includegraphics[width=0.95\textwidth]{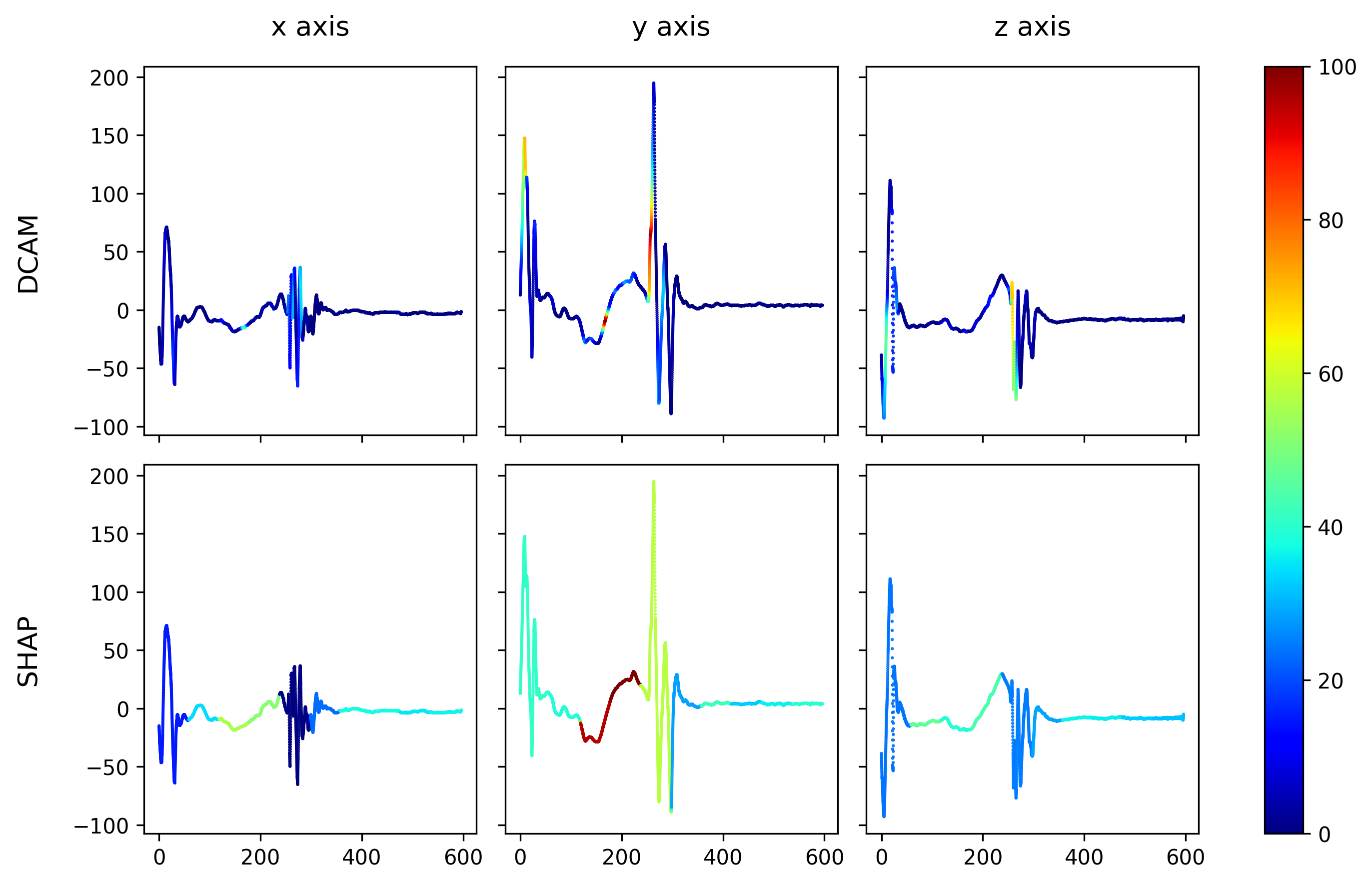}
    \caption{ Different saliency maps for the same CMJ instance correctly classified as Legs bending: both explanation methods focus on $y$ axis, but different parts. SHAP focuses on the small peak, around time step $200$ while dCAM focuses more on the beginning of the following higher peak.}
    \label{fig:CMJ_legs bending_appendix}
\end{figure}

\end{appendix}




\end{document}